 \pdfoutput=1
\documentclass[runningheads]{llncs}
\usepackage{graphicx}
\usepackage{comment}
\usepackage{amsmath,amssymb} 
\usepackage{color}
\usepackage[colorlinks]{hyperref}

\usepackage{booktabs}
\usepackage{multirow}
\graphicspath{ {./images/} }

\usepackage[utf8]{inputenc}
\usepackage{listings}
\usepackage{xcolor}

\usepackage{xcolor}

\newcommand{\mat}[1]{\ensuremath{\mathbf{#1}}}

\newcommand{\bx}{\mat{X}}
\newcommand{\bw}{\mat{W}}
\newcommand{\by}{\mat{Y}}
\newcommand{\mbR}{\mathbb{R}}
\newcommand{\ba}{\boldsymbol\alpha}

\newcommand{\cout}{C}

\begin{document}
\pagestyle{headings}
\mainmatter
\def\ECCVSubNumber{2461} 

\title{WeightNet: Revisiting the Design Space of Weight Networks} 

\titlerunning{WeightNet: Revisiting the Design Space of Weight Network}
%
\author{Ningning Ma\inst{1} \and
Xiangyu Zhang\inst{2}\thanks{Corresponding author} \and
Jiawei Huang\inst{2} \and
Jian Sun\inst{2} }
\authorrunning{Ningning Ma et al.}
%
\institute{Hong Kong University of Science and Technology\and MEGVII Technology\\
\email{nmaac@cse.ust.hk, \{zhangxiangyu,huangjiawei,sunjian\}@megvii.com}}
\maketitle

\begin{abstract}
We present a conceptually simple, flexible and effective framework for weight generating networks.
Our approach is general that unifies two current distinct and extremely effective SENet and CondConv into the same framework on weight space.
The method, called $WeightNet$, generalizes the two methods by simply adding one more grouped fully-connected layer to the attention activation layer.
We use the WeightNet, composed entirely of (grouped) fully-connected layers, to directly output the convolutional weight.
WeightNet is easy and memory-conserving to train, on the kernel space instead of the feature space.
Because of the flexibility, our method outperforms existing approaches on both ImageNet and COCO detection tasks, achieving better Accuracy-FLOPs and Accuracy-Parameter trade-offs.
The framework on the flexible weight space has the potential to further improve the performance.
Code is available at \url{https://github.com/megvii-model/WeightNet}. 

\end{abstract}

\section{Introduction}
Designing convolution weight is a key issue in convolution networks (CNNs). 
The weight-generating methods \cite{jia2016dynamic,ha2016hypernetworks,munkhdalai2017meta} using a network, which we call weight networks, provide an insightful neural architecture design space.  
These approaches are conceptually intuitive, easy and efficient to train.
Our goal in this work is to present a simple and effective framework, in the design space of weight networks, inspired by the rethinking of recent effective conditional networks.

 
Conditional networks (or dynamic networks)\cite{cao2019gcnet,hu2018squeeze,yang2019condconv}, which use extra sample-dependent modules to conditionally adjust the network, have achieved great success.
SENet \cite{hu2018squeeze}, an effective and robust attention module, helps many tasks achieve state-of-the-art results \cite{howard2019searching,tan2019mnasnet,tan2019efficientnet}.
Conditionally Parameterized Convolution (CondConv) \cite{yang2019condconv} uses over-parameterization to achieve great improvements but maintains the computational complexity at the inference phase.

Both of the methods consist of two steps: first, they obtain an attention activation vector, then using the vector, SE scales the feature channels, while CondConv performs a mixture of expert weights.
Despite they are usually treated as entirely distinct methods, they have some things in common. It is natural to ask: \textit{do they have any correlations?} 
We show that we can link the two extremely effective approaches, by generalizing them in the weight network space.

Our methods, called \textit{WeightNet}, extends the first step by simply adding one more layer for generating the convolutional weight directly (see Fig. \ref{fig:weightnet}).
The layer is a grouped fully-connected layer applied to the attention vector, generating the weight in a group-wise manner.
To achieve this, we rethink SENet and CondConv and discover that the subsequent operations after the first step can be cast to a \textit{grouped fully-connected layer}, however, they are particular cases.

In that grouped layer, the output is direct the convolution weight, but the input size and the group number are variable.
In CondConv the group number is discovered to be a minimum number of one and the input is small (4, 8, 16, etc.) to avoid the rapid growth of the model size.
In SENet the group is discovered to be the maximum number equal to the input channel number.

\begin{figure}[t]
\begin{center}
 \includegraphics[width=.9\textwidth]{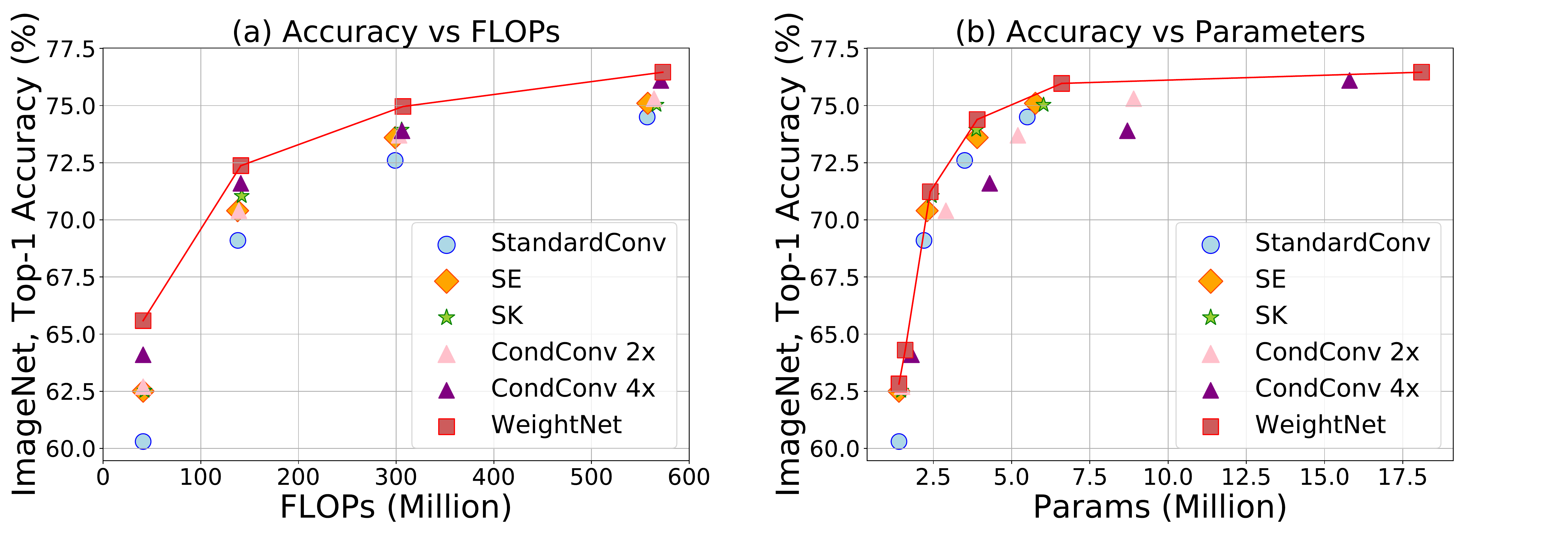}
\end{center}
\caption{\textbf{Accuracy vs. FLOPs vs. Parameters} comparisons on ImageNet, using ShuffleNetV2 \cite{ma2018shufflenet}. 
(a) The trade-off between accuracy and FLOPs;
(b) the trade-off between accuracy and number of parameters.}
\label{fig:tradeoff}
\end{figure}

Despite the two variants having seemingly minor differences, they have a large impact: 
they together control the parameter-FLOPs-accuracy tradeoff, leading to surprisingly different performance.
Intuitively, we introduce two hyperparameters $M$ and $G$, to control the input number and the group number, respectively.
The two hyperparameters have not been observed and investigated before, in the additional grouped fully-connected layer.
By simply adjusting them, we can strike a better trade-off between the representation capacity and the number of model parameters. 
We show by experiments on ImageNet classification and COCO detection the superiority of our method (Figure \ref{fig:tradeoff}). 


Our main contributions include: 1) 
First, we rethink the weight generating manners in SENet and CondConv, for the first time, to be complete fully-connected networks;
2) Second, only from this new perspective can we revisit the novel network design space in the weight space, which provides more effective structures than those in convolution design space (group-wise, point-wise, and depth-wise convolutions, etc). 
In this new and rarely explored weight space, there could be new structures besides fully-connected layers, there could also be more kinds of sparse matrix besides those in Fig. \ref{fig:group}. We believe this is a promising direction and hope it would have a broader impact on the vision community.


\section{Related Work}
\paragraph{\textbf{\emph{Weight generation networks}}}
Schmidhuber et al. \cite{schmidhuber1992learning} incorporate the "fast" weights into recurrent connections in RNN methods.
Dynamic filter networks \cite{jia2016dynamic} use filter-generating networks on video and stereo prediction.
HyperNetworks \cite{ha2016hypernetworks} decouple the neural networks according to the relationship in nature: a genotype (the hypernetwork), and a phenotype (the main network), that uses a small network to produce the weights for the main network, which reduces the number of parameters while achieving respectable results.
Meta networks \cite{munkhdalai2017meta} generate weights using a meta learner for rapid generalization across tasks.
The methods \cite{jia2016dynamic,ha2016hypernetworks,munkhdalai2017meta,platanios2018contextual} provide a worthy design space in the weight-generating network,
our method follows the spirits and uses a WeightNet to generate the weights.

\paragraph{\textbf{\emph{Conditional CNNs}}}
Different from standard CNNs \cite{simonyan2014very,he2016deep,szegedy2015going,zhang2018shufflenet,chollet2017xception,howard2017mobilenets,sandler2018mobilenetv2,howard2019searching},
conditional (or dynamic) CNNs \cite{lin2017runtime,liu2018dynamic,wu2018blockdrop,yu2018slimmable,keskin2018splinenets} use dynamic kernels, widths, or depths conditioned on the input samples, showing great improvement.
Spatial Transform Networks \cite{jaderberg2015spatial} learns to transform to warp the feature map in a parametric way.
Yang et al. \cite{yang2019condconv} proposed conditional parameterized convolution to mix the experts voted by each sample's feature.
The methods are extremely effective because they improve the Top-1 accuracy by more than 5\% on the ImageNet dataset, which is a great improvement.
Different from dynamic features or dynamic kernels, another series of work \cite{wang2018skipnet,huang2017multi} focus on dynamic depths of the convolutional networks, that skip some layers for different samples.

\paragraph{\textbf{\emph{Attention and gating mechanism}}}
Attention mechanism \cite{vaswani2017attention,luong2015effective,bahdanau2014neural,wang2017residual,woo2018cbam} is also a kind of conditional network, that adjusts the networks dependent on the input.
Recently the attention mechanism has shown its great improvement.
Hu et al. \cite{hu2018squeeze} proposed a block-wise gating mechanism to enhance the representation ability, where they adopted a squeeze and excitation method to use global information and capture channel-wise dependencies.
SENet achieves great success by not only winning the ImageNet challenge \cite{deng2009imagenet}, but also helping many structures to achieve state-of-the-art performance \cite{howard2019searching,tan2019mnasnet,tan2019efficientnet}.
In GaterNet \cite{chen2018gaternet}, a gater network was used to predict binary masks for the backbone CNN, which can result in performance improvement.
Besides, Li et al. \cite{li2019selective} introduced a kernel selecting module, where they added attention to kernels with different sizes to enhance CNN's learning capability. 
In contrast, WeightNet is designed on kernel space which is more time-conserving and memory-conserving than feature space.

\section{WeightNet}
\label{sec:method}

\begin{figure}[t]
\begin{center}
 \includegraphics[width=0.9\textwidth]{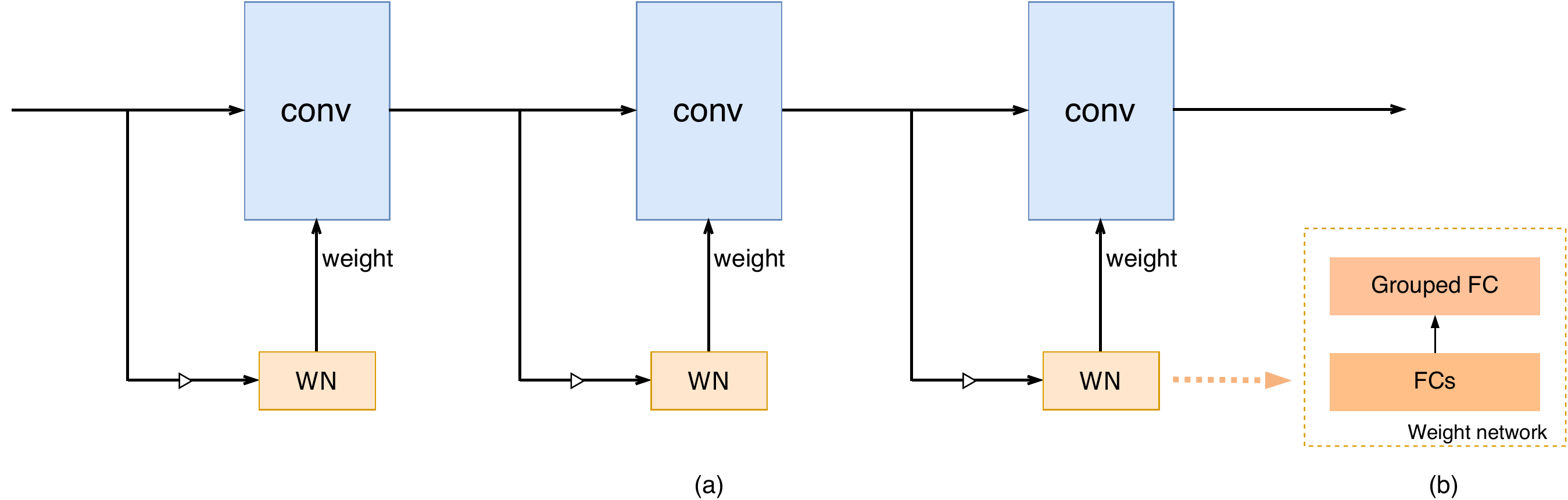}
\end{center}
\caption{\textbf{The WeightNet structure.}
The convolutional weight is generated by WeightNet that is comprised entirely of (grouped) fully-connected layers.
The symbol ($\rhd$) represents the dimension reduction (global average pool) from feature space ($C \times H \times W$) to kernel space ($C$).
The 'FC' denotes a fully-connected layer, 'conv' denotes convolution, and 'WN' denotes the WeightNet.}
\label{fig:weightnet}
\end{figure}

The WeightNet generalizes the current two extremely effective modules in weight space.
Our method is conceptually simple: both SENet and CondConv generate the $activation\ vector$ using a global average pooling (GAP) and one or two fully-connected layers followed with a non-linear sigmoid operation; to this we simply add one more grouped fully-connected layer, to generate the weight $directly$ (Fig. \ref{fig:weightnet}).
This is different from common practice that applies the vector to feature space and we avoid the memory-consuming training period.

WeightNet is computationally efficient because of the dimension reduction (GAP) from $C\times H\times W$ dimension to a 1-D dimension $C$.
Evidently, the WeightNet only consists of (grouped) fully-connected layers.
We begin by introducing the matrix multiplication behaviors of (grouped) fully-connected operations.

\subsubsection{Grouped fully-connected operation}
\begin{figure}[t]
\begin{center}
 \includegraphics[width=0.44\textwidth]{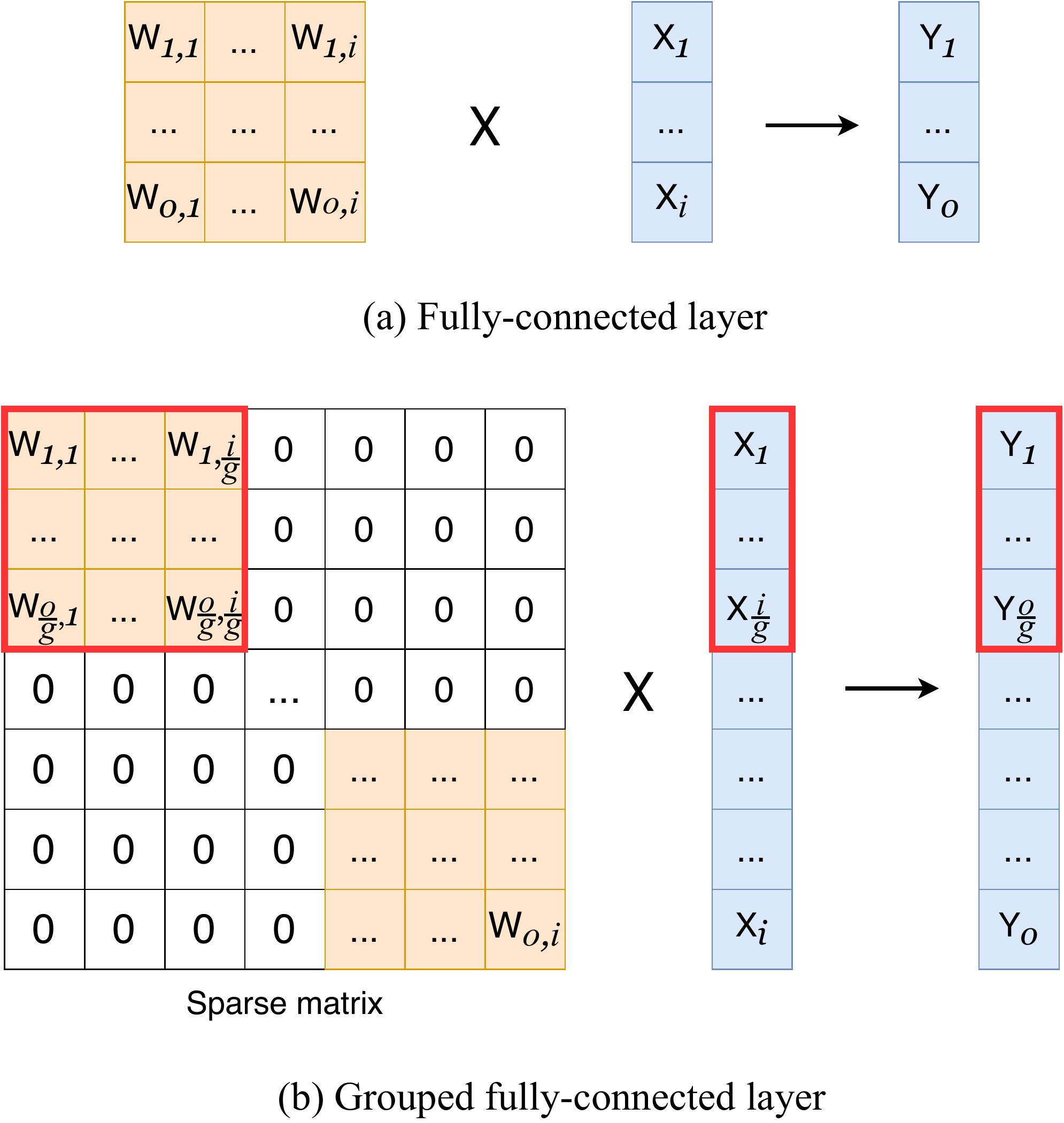}
\end{center}
\caption{The \textbf{matrix multiplication} behaviors of the (grouped) fully-connected operations.
Here $i$, $o$ and $g$ denote the numbers of the input channel, output channel and group number.
(a) A standard matrix multiplication representing a fully-connected layer.
(b) With the weight in a block diagonal sparse matrix, it becomes a general grouped fully-connected layer. Each group (red box) is exactly a standard matrix multiplication in (a), with $i/g$ input channels and $o/g$ output channels.
Fig. (a) is a special case of Fig. (b) where $g=1$.  
}
\label{fig:matrix}
\end{figure}

Conceptually, neurons in a fully-connected layer have full connections and thus can be computed with a matrix multiplication, in the form $Y=WX$ (see Fig. \ref{fig:matrix} (a)).
Further, neurons in a grouped fully-connected layer have group-wise sparse connections with activations in the previous layer.

Formally, in Fig. \ref{fig:matrix} (b), the neurons are divided exactly into $g$ groups, each group (with $i/g$ inputs and $o/g$ outputs) performs a fully-connected operation (see the red box for example).
One notable property of this operation, which can be easily seen in the graphic illustration, is that the weight matrix becomes a sparse, $block\ diagonal\ matrix$, with a size of $(o/g \times i/g)$ in each block.

Grouped fully-connected operation is a general form of fully-connected operation where the group number is one.
Next, we show how it generalizes CondConv and SENet: use the grouped fully-connected layer to replace the subsequent operations after the activation vector and directly output the generated weight.

\subsubsection{Denotation}

We denote a convolution operation with the input feature map $\bx\in\mbR^{\cout\times h\times w}$, the output feature map $\by \in\mbR^{\cout \times h'\times w' }$, and the convolution weight $\bw'\in\mbR^{ \cout\times \cout\times k_h\times k_w}$.
For simplicity, but without loss of generality, it is assumed that the number of the input channels equals to that of output channels, here $(h,w),(h',w'),(k_h,k_w)$ denote the 2-D heights and the widths for the input, output, and kernel.
Therefore, we denote the convolution operation using the symbol $(*)$: $\by_c = \bw'_c * \bx$.
We use $\ba$ to denote the attention activation vector in CondConv and SENet.

\subsection{Rethinking CondConv}
Conditionally parameterized convolution (CondConv) \cite{yang2019condconv} is a mixture of $m$ experts of weights, voted by a $m$-dimensional vector $\ba$, that is sample-dependent and makes each sample's weight dynamic.

Formally, we begin with reviewing the first step in CondConv, it gets $\ba$ by a global average pooling and a fully-connected layer $\bw_{fc1}$, followed by a sigmoid $\sigma(\cdot)$ : $\ba = \sigma(\bw_{fc1} \times \frac{1}{hw} \sum_{i \in h, j \in w}{\bx_{c,i,j}})$, here $(\times)$ denotes the matrix multiplication, $\bw_{fc1}\in\mbR^{ m\times C}$, $\alpha\in\mbR^{ m\times 1}$.

Next, we show the following mixture of expert operations in the original paper can essentially be replaced by a fully-connected layer.
The weight is generated by multiple weights:
$\bw'=\alpha_1 \cdot \bw_1 + \alpha_2 \cdot \bw_2 +...+ \alpha_m \cdot \bw_m$, here $\bw_i \in\mbR^{ \cout\times \cout\times k_h\times k_w}, (i\in \{1,2,...,m\})$.
We rethink it as follows: 

\begin{equation}
\label{eqa:condconv}
\begin{split}
 & \bw'=\bw^T \times \ba \\
where \ & \bw = [\bw_1 \bw_2 ... \bw_m]
 \end{split}
\end{equation}

Here $\bw\in\mbR^{ m\times \cout \cout k_h k_w}$ denotes the matrix concatenation result, $(\times)$ denotes the matrix multiplication (fully-connected in Fig. \ref{fig:matrix}a).
Therefore, the weight is generated by simply adding one more layer $(\bw)$ to the activation layer.
That layer is a fully-connected layer with $m$ inputs and $C\times C\times k_h\times k_w$ outputs.

This is different from the practice in the original paper in the training phase.
In that case, it is memory-consuming and suffers from the batch problem when increasing $m$ (batch size should be set to one when $m>4$).
In this case, we train with large batch sizes efficiently.

\subsection{Rethinking SENet}
Squeeze and Excitation (SE) \cite{hu2018squeeze} block is an extremely effective "plug-n-play" module that is acted on the feature map.
We integrate the SE module into the convolution kernels and discover it can also be represented by adding one more grouped fully-connected layer to the activation vector $\ba$.
We start from the reviewing of the $\ba$ generation process.
It has a similar process with CondConv: a global average pool, two fully-connected layer with non-linear ReLU $(\delta)$ and sigmoid $(\sigma)$: $\ba = \sigma(\bw_{fc_2} \times \delta( \bw_{fc1} \times \frac{1}{hw} \sum_{i \in h, j \in w}{\bx_{c,i,j}}))$, here $\bw_{fc1}\in\mbR^{ C/r \times C}$, $\bw_{fc2}\in\mbR^{ C\times C/r}$, $(\times)$ in the equation denotes the matrix multiplication.
The two fully-connected layers here are mainly used to reduce the number of parameters because $\ba$ here is a $C$-dimensional vector, a single layer is parameter-consuming. 

Next, in common practice the block is used before or after a convolution layer, $\ba$ is computed right before a convolution (on the input feature $\bx$): $\by_c = \bw'_c * (\bx \cdot \ba)$, or right after a convolution (on the output feature $\by$): $\by_c = (\bw'_c * \bx) \cdot \ba_c$, here $(\cdot)$ denotes dot multiplication broadcasted along the $C$ axis. 
In contrast, on kernel level, we analyze the case that SE is acted on $\bw'$: $\by_c = (\bw'_c \cdot \ba_c)* \bx $.
Therefore we rewrite the weight to be $\bw' \cdot \ba$, the $(\cdot)$ here is different from the $(\times)$ in Equ. \ref{eqa:condconv}. In that case, a dimension reduction is performed; in this case, no dimension reduction.
Therefore, it is essentially a grouped sparse connected operation, that is a particular case of Fig. \ref{fig:matrix} (b), with $C$ inputs, $C\times C\times k_h\times k_w$ outputs, and $C$ groups.

\begin{table}[t]
\begin{center}
\caption{ \textbf{Summary of the configure} in the grouped fully-connected layer. $\lambda$ is the proportion of input size to group number, representing the major increased parameters.}
\label{table:rethink}
\resizebox{0.7\textwidth}{!}{
\begin{tabular}{lcccc}
\toprule[1.0pt]
Model & Input size & Group number & $\lambda$ &Output size		 \\ \hline
CondConv & 	m & 1 &m& $C\times C \times k_h \times k_w$ \\  
SENet & 	C & C&1 & $C\times C \times k_h \times k_w$ \\  
WeightNet & 	$M\times C$ & $G\times C$ &M/G& $C\times C \times k_h \times k_w$ \\  
 \toprule[1.0pt]
\end{tabular}
}
\end{center}
\end{table}

\subsection{WeightNet Structure}
\label{sec:dynconv}
\begin{figure}[t]
\begin{center}
 \includegraphics[width=0.5\textwidth]{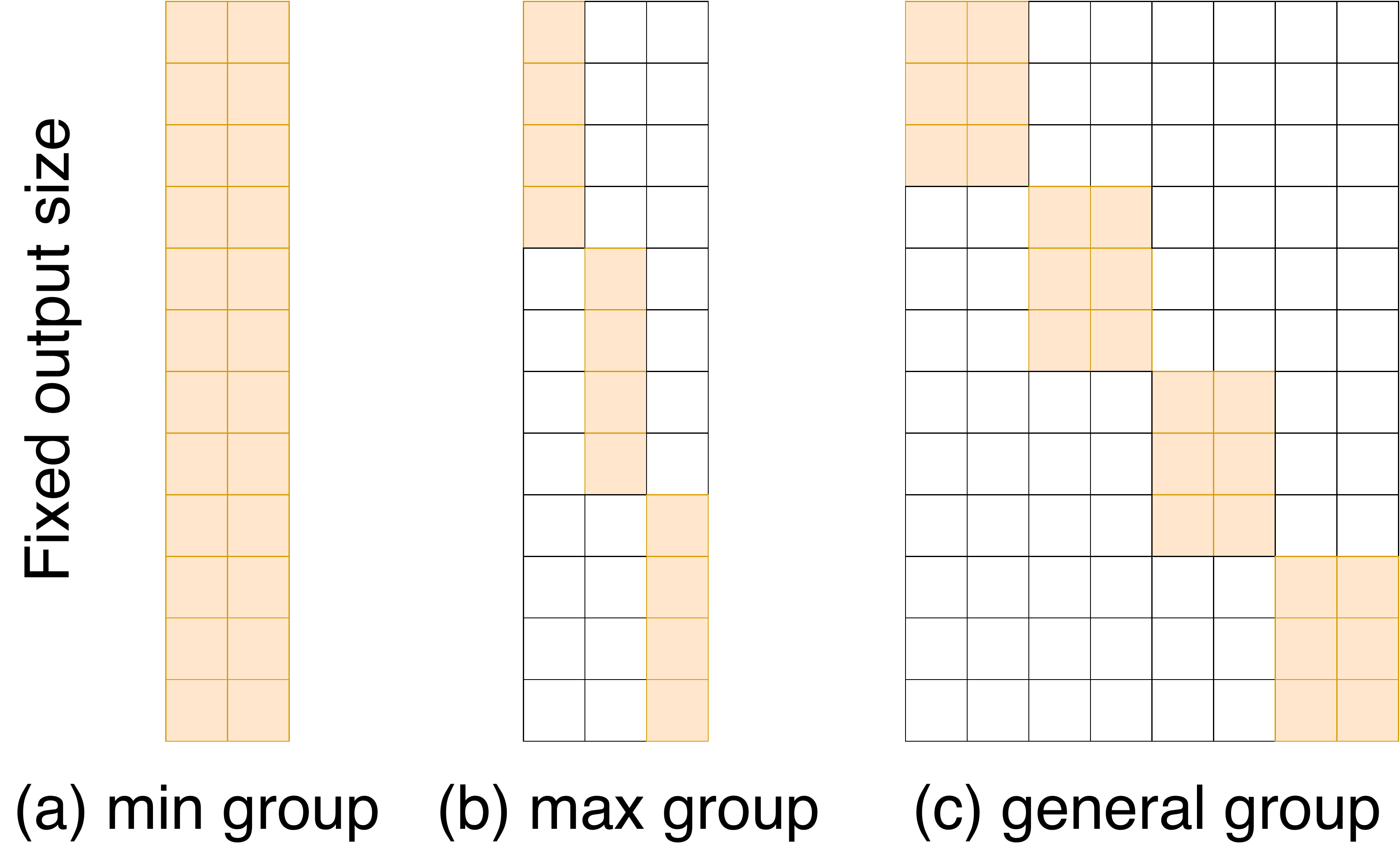}
\end{center}
\caption{The diagrams of the different cases in the \textbf{block diagonal matrix} (Fig. \ref{fig:matrix} b), that can represent the weights of the grouped fully-connected layer in CondConv, SENet and the general WeightNet.
They output the same fixed size (convolution kernel's size $C\times C\times k_h\times k_w$), but have different group numbers:
(a) the group number has a minimum number of one (CondConv), (b) the group number has a maximum number equals to the input size $C$ (SENet), since (a) and (b) are extreme cases, (c) shows the general group number between 1 and the input size (WeightNet).
}
\label{fig:group}
\end{figure}

By far, we note that the group number in the general grouped fully-connected layer (Fig. \ref{fig:matrix} b) has values range from 1 to the channel number.
That is, the group has a minimum number of one and has a maximum number of the input channel numbers.
It, therefore, generalizes the CondConv, where the group number takes the minimum value (one), and the SENet, where it takes the maximum value (the input channel number). 
We conclude that they are two extreme cases of the general grouped fully-connected layer (Fig. \ref{fig:group}).

We summarize the configure in the grouped fully-connected layer (in Table \ref{table:rethink}) and generalize them using two additional hyperparameters $M$ and $G$.
To make the group number more flexible, we set it by combining the channel $C$ and a constant hyperparameter $G$,
Moreover, another hyperparameter $M$ is used to control the input number, thus $M$ and $G$ together to control the parameter-accuracy tradeoff.
The layer in CondConv is a special case with $M=m/C, G=1/C$, while for SENet $M=1,G=1$.
We constrain $M\times C$ and $G\times C$ to be integers, $M$ is divisible by $G$ in this case.
It is notable that the two hyperparameters are right there but have not been noticed and investigated.

%
%
%
%
%

\subsubsection{Implementation details}

For the activation vector $\ba$'s generating step, since $\ba$ is a ($M\times C$)-dimensional vector, it may be large and parameter-consuming, therefore, we use two fully-connected layers with a reduction ratio $r$.
It has a similar process with the two methods: a global average pool, two fully-connected layer with non-linear sigmoid $(\sigma)$: $\ba = \sigma(\bw_{fc_2} \times \bw_{fc1} \times \frac{1}{hw} \sum_{i \in h, j \in w}{\bx_{c,i,j}})$, here $\bw_{fc1}\in\mbR^{ C/r\times C}$, $\bw_{fc2}\in\mbR^{ MC\times C/r}$, $(\times)$ denotes the matrix multiplication, $r$ has a default setting of 16.

In the second step, we adopt a grouped fully-connected layer with $M\times C$ input, $C\times C \times k_h \times k_w$ output, and $G\times C$ groups.
We note that the structure is a straightforward design, and more complex structures have the potential to improve the performance further, but it is beyond the focus of this work.

\subsubsection{Complexity analysis}
The structure of WeightNet decouples the convolution computation and the weight computation into two separate branches (see Fig. \ref{fig:weightnet}).
Because the spatial dimensions ($h\times w$) are reduced before feeding into the weight branch,
the computational amount (FLOPs) is mainly in the convolution branch.
The FLOPs complexities in the convolution and weight branches are $O(hwCCk_hk_w)$ and $O(MCC k_h k_w/G)$, the latter is relatively negligible. 
The parameter complexities for each branch are zero and $O(M/G\times C\times C \times k_h \times k_w)$, which is $M/G$ times of normal convolution.
We notate $\lambda$ to represent it (Table \ref{table:rethink}).

\subsubsection{Training with batch dimension}
The weight generated by WeightNet has a dimension of batch size,
here we briefly introduce the training method related to the batch dimension. 
We denote $B$ as batch size and reshape the input $\bx$ of the convolution layer to $(1, B \times \cout, h, w)$. Thus $\bx$ has $B \times \cout$ channel numbers, which means we regard different samples in the same batch as different channels.
Next, we reshape the generated weight $\bw$ to $(B, \cout, \cout, k_h, k_w)$.
Then it becomes a group convolution, with a group number of $B$, the inputs and the outputs in each group are both equal to $\cout$.
Therefore, we use the same memory-conserving method for both training and inference periods, and this is different from CondConv.

\section{Experiments}


In this section, we evaluate the WeightNet on classification and COCO detection tasks \cite{lin2014microsoft}.
In classification task, we conduct experiments on a light-weight CNN model ShuffleNetV2 \cite{ma2018shufflenet} and a deep model ResNet50 \cite{he2016deep}. 
In the detection task, we evaluate our method's performance on distinct backbone models under RetinaNet.
In the final analysis, we conduct ablation studies and investigate the properties of WeightNet in various aspects.

\subsection{Classification}

We conduct image classification experiments on ImageNet 2012 classification dataset, which includes 1000 classes \cite{russakovsky2015imagenet}. Our models are first trained on the training dataset that consists of 1.28 million images and then evaluated over 50k images in the validation dataset. 
For the training settings, all the ShuffleNetV2 \cite{ma2018shufflenet} models are trained with the same settings as \cite{ma2018shufflenet}. For ResNet-50, we use a linear decay scheduled learning rate starting with 0.1, a batch size of 256, a weight decay of 1e-4, and 600k iterations. 

\subsubsection{ShuffleNetV2}
To investigate the performance of our method on light-weight convolution networks, we construct experiments based on a recent effective network ShuffleNetV2 \cite{ma2018shufflenet}.
For a fair comparison, we retrain all the models by ourselves, using the same code base.
We replace the standard convolution kernels in each bottleneck with our proposed WeightNet, and control FLOPs and the number of parameters for fairness comparison.

As shown in Table \ref{table:rethink}, $\lambda$ is utilized to control the number of parameters in a convolution.
For simplicity, we fix $G=2$ when adjusting $\lambda$.
In our experiments, $\lambda$ has several sizes $\{1\times,2\times,4\times,8\times\}$.
To make the number of channels conveniently divisible by $G$ when scaling the number of parameters, we slightly adjust the number of channels for ShuffleNetV2 1$\times$ and $2\times$. 

We evaluate the WeightNet from two aspects. 
Table \ref{table:snetv2} reports the performance of our method considering parameters.
The experiments illustrate that our method has significant advantages over the other counterparts under the same FLOPs and the same number of parameter constraints.
ShuffleNetV2 0.5$\times$ gains 3\% Top-1 accuracy without additional computation budgets.

In Table \ref{table:snetv2flops}, we report the advantages after applying our method on ShuffleNetV2 with different sizes. Considering in practice, the storage space is sufficient. Therefore, without the loss of fairness, we only constrain the Flops to be the same and tolerate the increment of parameters.

ShuffleNet V2 (0.5$\times$) gains 5.7\% Top-1 accuracy which shows further significant improvements by adding a minority of parameters.
ShuffleNet V2 (2$\times$) gains 2.0\% Top-1 accuracy.

\begin{table}[t]
\begin{minipage}{0.49\linewidth}
\centering

\caption{\textbf{ImageNet classification} results of the WeightNet on ShuffleNetV2 \cite{ma2018shufflenet}. For fair comparison, we control the values of $\lambda$ to be 1$\times$, to make sure that the experiments are under the same FLOPs and the same number of parameters.}
\label{table:snetv2}
\resizebox{1\textwidth}{!}{
\begin{tabular}{lccc}
\toprule[1.0pt]
 Model & \# Params & FLOPs & Top-1 err. \\ \hline 
ShuffleNetV2 (0.5$\times$) 			& 1.4M & 41M & 39.7 \\ 
+ WeightNet (1$\times$) 	& 1.5M & 41M & \textbf{36.7} \\ \hline

ShuffleNetV2 (1$\times$) 			& 2.2M & 138M & 30.9 \\ 
+ WeightNet (1$\times$) 	& 2.4M & 139M & \textbf{28.8} \\ \hline

ShuffleNetV2 (1.5$\times$) 			& 3.5M & 299M & 27.4 \\ 
+ WeightNet (1$\times$) 	& 3.9M & 301M & \textbf{25.6} \\ \hline

ShuffleNetV2 (2$\times$) 			& 5.5M & 557M & 25.5 \\ 
+ WeightNet (1$\times$) 	& 6.1M & 562M& \textbf{24.1} \\ \toprule[1.0pt]
\end{tabular}
}

\end{minipage}
\hfill
\begin{minipage}{0.49\linewidth}  
\centering

\caption{\textbf{ImageNet classification} results of the WeightNet on ShuffleNetV2 \cite{ma2018shufflenet}. The comparison is under the same FLOPs and regardless of the number of parameters. 
To obtain the optimum performance, we set the $\lambda$ to \{8$\times$, 4$\times$, 4$\times$, 4$\times$\} respectively.
}
\label{table:snetv2flops}
\resizebox{1\textwidth}{!}{
\begin{tabular}{lccc}
\toprule[1.0pt]
 Model & \# Params & FLOPs & Top-1 err. \\ \hline 
ShuffleNetV2 (0.5$\times$) 			& 1.4M & 41M & 39.7 \\ 
+ WeightNet (8$\times$) 	& 2.7M & 42M & \textbf{34.0} \\ \hline

ShuffleNetV2 (1$\times$) 			& 2.2M & 138M & 30.9 \\ 
+ WeightNet (4$\times$) 	& 5.1M & 141M & \textbf{27.6} \\ \hline

ShuffleNetV2 (1.5$\times$) 			& 3.5M & 299M & 27.4 \\ 
+ WeightNet (4$\times$) 	& 9.6M & 307M & \textbf{25.0} \\ \hline

ShuffleNetV2 (2$\times$) 			& 5.5M & 557M & 25.5 \\ 
+ WeightNet (4$\times$) 	& 18.1M & 573M & \textbf{23.5} \\ \toprule[1.0pt]
\end{tabular}
}

\end{minipage}
\end{table}

To further investigate the improvement of our method, we compare our method with some recent effective conditional CNN methods under the same FLOPs and the same number of parameters.
For the network settings of CondConv \cite{yang2019condconv}, we replace standard convolutions in the bottlenecks with CondConv, and change the number of experts as described in CondConv to adjust parameters, as the number of experts grows, the number of parameters grows.
To reveal the model capacity under the same number of parameters, for our proposed WeightNet, we control the number of parameters by changing $\lambda$.
Table \ref{table:sescc} describes the comparison between our method and other counterpart effective methods, from which we observe our method outperforms the other conditional CNN methods under the same budgets.
The Accuracy-Parameters tradeoff and the Accuracy-FLOPs tradeoff are shown in Figure \ref{fig:tradeoff}.

From the results, we can see SE and CondConv boost the base models of all sizes significantly.
However, CondConv has major improvements in smaller sizes especially, but as the model becomes larger, the smaller the advantage it has.
For example, CondConv performs better than SE on ShuffleNetV2 0.5$\times$ but SE performs better on ShuffleNetV2 2$\times$.
In contrast, we find our method can be uniformly better than SE and CondConv.

To reduce the overfitting problem while increasing parameters, we add dropout \cite{hinton2012improving} for models with more than 3M parameters.
As we described in Section \ref{sec:dynconv}, $\lambda$ represents the increase of parameters, so we measure the capacity of networks by changing parameter multiplier $\lambda$ in $\{1\times, 2\times, 4\times, 8\times \}$. 
We further analyze the effect of $\lambda$ and the grouping hyperparameter $G$ on each filter in the ablation study section.

\begin{table}[t]
\begin{center}
\caption{\textbf{Comparison with recently effective attention modules} on ShuffleNetV2 \cite{ma2018shufflenet} and ResNet50 \cite{he2016deep}. We show results on ImageNet.}
\label{table:sescc}
\resizebox{.6\textwidth}{!}{
\begin{tabular}{lcccc}
\toprule[1.0pt]
 Model & \# Params & FLOPs & Top-1 err. \\ \hline 
 ShuffleNetV2 \cite{ma2018shufflenet} (0.5$\times$) 			& 1.4M & 41M & 39.7 \\ 
+ SE \cite{hu2018squeeze}		& 1.4M & 41M & 37.5\\ 
+ SK \cite{li2019selective}		& 1.5M & 42M & 37.5\\ 
 + CondConv \cite{yang2019condconv} (2$\times$)	& 1.5M& 41M& 37.3\\ 
 + WeightNet (1$\times$) & 1.5M & 41M & \textbf{36.7} \\ 
 + CondConv \cite{yang2019condconv} (4$\times$)	& 1.8M& 41M& 35.9 \\ 
 + WeightNet (2$\times$) & 1.8M& 41M& \textbf{35.5}\\ \hline 

ShuffleNetV2 \cite{ma2018shufflenet} (1.5$\times$) 			& 3.5M & 299M & 27.4 \\ 
 + SE \cite{hu2018squeeze}		& 3.9M & 299M & 26.4 \\ 
 + SK \cite{li2019selective}		& 3.9M & 306M & 26.1 \\ 
 + CondConv \cite{yang2019condconv} (2$\times$)	& 5.2M & 303M & 26.3 \\ 
 + WeightNet (1$\times$) & \textbf{3.9M} & 301M & \textbf{25.6} \\ 
 + CondConv \cite{yang2019condconv} (4$\times$)	& 8.7M & 306M & 26.1 \\ 
 + WeightNet (2$\times$)	& \textbf{5.9M} & \textbf{303M} & \textbf{25.2} \\ \hline 


ShuffleNetV2 \cite{ma2018shufflenet} (2.0$\times$) 			& 5.5M & 557M & 25.5 \\ 
 + WeightNet (2$\times$) & 10.1M & 565M & \textbf{23.7} \\ \hline
 	
ResNet50 \cite{he2016deep} 			& 25.5M & 3.86G & 24.0 \\ 
+ SE \cite{hu2018squeeze} 	& 26.7M & 3.86G & 22.8 \\  
+ CondConv \cite{yang2019condconv} (2$\times$) 	& 72.4M & 3.90G & 23.4 \\  
+ WeightNet (1$\times$) 	&31.1M & 3.89G & \textbf{22.5} \\ \toprule[1.0pt]

\end{tabular}
}
\end{center}
\end{table}

\subsubsection{ResNet50}


For larger classification models, we conduct experiments on ResNet50 \cite{he2016deep}.
We use a similar way to replace the standard convolution kernels in ResNet50 bottlenecks with our proposed WeightNet.
Besides, we train the conditional CNNs utilizing the same training settings with the base ResNet50 network.

In Table \ref{table:sescc}, based on ResNet50 model, we compare our method with SE \cite{hu2018squeeze} and CondConv \cite{yang2019condconv} under the same computational budgets.
It's shown that our method still performs better than other conditional convolution modules. 
We perform CondConv (2$\times$) on ResNet50, the results reveal that it does not have further improvement comparing with SE, although CondConv has a larger number of parameters. 
We conduct our method (1$\times$) by adding limited parameters and it also shows further improvement comparing with SE.
Moreover, ShuffleNetV2 \cite{ma2018shufflenet} (2$\times$) with our method performs better than ResNet50, with only 40\% parameters and 14.6\% FLOPs.

\subsection{Object Detection}
We evaluate the performance of our method on COCO detection \cite{lin2014microsoft} task.
The COCO dataset has 80 object categories.
We use the $trainval35k$ set for training and use the $minival$ set for testing.
For a fair comparison, we train all the models with the same settings.
The batch size is set to 2, the weight decay is set to 1e-4 and the momentum is set to 0.9.
We use anchors for 3 scales and 3 aspect ratios and use a 600-pixel train and test image scale.
We conduct experiments on RetinaNet \cite{lin2017focal} using ShuffleNetV2 \cite{ma2018shufflenet} as the backbone feature extractor.
We compare the backbone models of our method with the standard CNN models. 

Table \ref{table:det} illustrates the improvement of our method over standard convolution on the RetinaNet framework.
For simplicity we set $G=1$ and adjust the size of WeightNet to $\{2\times, 4\times\}$.
As we can see our method improves the mAP significantly by adding a minority of parameters.
ShuffleNetV2 (0.5$\times$) with WeightNet (4$\times$) improves 4.6 mAP by adding few parameters under the same FLOPs.

%
%
%

To compare the performance between WeightNet and CondConv \cite{yang2019condconv} under the same parameters, we utilize ShuffleNetV2 0.5$\times$ as the backbone and investigate the performances of all CondConv sizes.
Table \ref{table:det_abl} reveals the clear advantage of our method over CondConv.
Our method outperforms CondConv uniformly under the same computational budgets.
As a result, our method is indeed robust and fundamental on different tasks.
%


\begin{table}[t]
\begin{minipage}{0.49\linewidth}
\centering

\caption{\textbf{Object detection} results comparing with baseline backbone. We show RetinaNet \cite{lin2017focal} results on COCO.}
\label{table:det}
\resizebox{1\textwidth}{!}{
\begin{tabular}{lcccc}
\toprule[1.0pt]
 Backbone & \# Params & FLOPs & mAP \\ \hline 
ShuffleNetV2 \cite{ma2018shufflenet} (0.5$\times$) 			& 1.4M & 41M & 22.5 \\ 
 + WeightNet (4$\times$) 	& 2.0M & 41M & \textbf{27.1} \\ \hline

ShuffleNetV2 \cite{ma2018shufflenet} (1.0$\times$) 			& 2.2M & 138M & 29.2 \\ 
 + WeightNet (4$\times$) 	& 4.8M & 141M & \textbf{32.1}\\ \hline

ShuffleNetV2 \cite{ma2018shufflenet} (1.5$\times$) 			& 3.5M & 299M & 30.8 \\ 
 + WeightNet (2$\times$)	& 5.7M & 303M & \textbf{33.3} \\ \hline

ShuffleNetV2 \cite{ma2018shufflenet} (2.0$\times$) 			& 5.5M & 557M & 33.0 \\ 
 + WeightNet (2$\times$)	& 9.7M& 565M & \textbf{34.0} \\ \toprule[1.0pt]
\end{tabular}
}

\end{minipage}
\hfill
\begin{minipage}{0.49\linewidth}  
\centering

\caption{\textbf{Object detection} results comparing with other conditional CNN backbones. We show RetinaNet \cite{lin2017focal} results on COCO.}
\label{table:det_abl}
\resizebox{1\textwidth}{!}{
\begin{tabular}{lccc}
\toprule[1.0pt]
 Backbone & \# Params & FLOPs & mAP \\ \hline 
ShuffleNetV2 \cite{ma2018shufflenet} (0.5$\times$) 			& 1.4M & 41M & 22.5 \\ 
+ SE \cite{hu2018squeeze}		& 1.4M & 41M & 25.0\\ 
+ SK \cite{li2019selective}		& 1.5M & 42M & 24.5\\ 
+ CondConv \cite{yang2019condconv} (2$\times$) & 1.5M & 41M & 25.8 \\
+ CondConv \cite{yang2019condconv} (4$\times$) & 1.8M & 41M & 25.0 \\
+ CondConv \cite{yang2019condconv} (8$\times$) & 2.3M & 42M & 26.4 \\
+ WeightNet (4$\times$) 	& 2.0M & 41M & \textbf{27.1} \\ 

\toprule[1.0pt]
\end{tabular}
}

\end{minipage}
\end{table}

\subsection{Ablation Study and Analysis}


\begin{table}[t]
\begin{minipage}{0.49\linewidth}
\centering

\caption{\textbf{Ablation on $\lambda$.} The table shows the ImageNet Top-1 err. results. The experiments are conducted on ShuffleNetV2 \cite{ma2018shufflenet}. By increasing $\lambda$ in the range \{1,2,4,8\}, the FLOPs does not change and the number of parameters increases. }
\label{table:lambda}
\resizebox{1\textwidth}{!}{
\begin{tabular}{lcccc}
\toprule[1.0pt]
 & \multicolumn{4}{c}{$\lambda$} \\ \hline 
Model &1 & 2 & 4 & 8			 \\ \hline
ShuffleNetV2 (0.5$\times$) & 	36.7 & 35.5 & 34.4 & \textbf{34.0} \\  
ShuffleNetV2 (1.0$\times$) & 	28.8 & 28.1 & \textbf{27.6} & 27.8 \\  
ShuffleNetV2 (1.5$\times$) & 	25.6 & 25.2 & \textbf{25.0} & 25.3 \\  
ShuffleNetV2 (2.0$\times$) & 	24.1 & 23.7 & \textbf{23.5} & 24.0\\ \toprule[1.0pt]
\end{tabular}
}

\end{minipage}
\hfill
\begin{minipage}{0.49\linewidth}  
\centering

\caption{\textbf{Ablation on $G$.} We tune the group hyperparameter $G$ to \{1,2,4\}, we keep $\lambda=1$. The results are ImageNet Top-1 err. The experiments are conducted on ShuffleNetV2 \cite{ma2018shufflenet}.}
\label{table:group}
\resizebox{1\textwidth}{!}{
\begin{tabular}{lcccc}
\toprule[1.0pt]
 Model & G & \# Params & FLOPs & Top-1 err. \\ \hline  
\multicolumn{1}{c}{\multirow{4}{*}{\begin{tabular}[c]{@{}c@{}}ShuffleNetV2 \\(0.5$\times$)\end{tabular}}} &G=1 			& 1.4M & 41M & 37.18 \\ 
&G=2 			& 1.5M & 41M & 36.73 \\ 
&G=4	   	& 1.5M & 41M & \textbf{36.37} \\ \hline

\multicolumn{1}{c}{\multirow{4}{*}{\begin{tabular}[c]{@{}c@{}}ShuffleNetV2 \\(1.0$\times$)\end{tabular}}} &G=1 			& 2.3M & 139M & 29.09 \\ 
&G=2 			& 2.4M & 139M & 28.77 \\ 
&G=4	   	& 2.6M & 139M & \textbf{28.76} \\ 

\toprule[1.0pt]
\end{tabular}
}

\end{minipage}
\end{table}


\begin{table}[t]
\begin{minipage}{0.48\linewidth}
\centering

\caption{Ablation study on \textbf{different stages}.
The $\checkmark$ means the convolutions in that stage is integrated with our proposed WeightNet. 
}
\label{table:stage}
\begin{tabular}{cccc}
\\ \toprule[1.0pt] 
Stage2 & Stage3 & Stage4 & Top-1 err.  \\ \hline
 $\checkmark$ &  &  &  39.13  \\
  &  $\checkmark$ &  &  36.82  \\
  &  & $\checkmark$ &  36.43  \\
 $\checkmark$ & $\checkmark$ &  &  35.73  \\
 $\checkmark$ &  &  $\checkmark$ &  36.44 \\
   &  $\checkmark$ & $\checkmark$ &   \textbf{35.30}  \\
 $\checkmark$ &  $\checkmark$ & $\checkmark$ &  35.47   \\
\toprule[1.0pt]
\end{tabular}
\end{minipage}
\hfill
\begin{minipage}{0.48\linewidth}  
\centering

\caption{ Ablation study on the number of the \textbf{global average pooling} operators in the whole network. We conduct ShuffleNetV2 0.5$\times$ experiments on ImageNet dataset. 
We compare the cases: one global average pooling in 1) each stage, 2) each block, and 3) each layer.
GAP represents global average pooling in this table.}
\label{table:gap}
\begin{tabular}{lc}\\
\toprule[1.0pt]
 & Top-1 err. \\ \hline 
 Stage wise GAP		& 37.01 \\ 
 Block wise GAP	& 35.47 \\ 
 Layer wise GAP & 35.04 \\ 
\toprule[1.0pt]
\end{tabular}
\end{minipage}
\end{table}

\subsubsection{The influence of $\lambda$.}
By tuning $\lambda$, we control the number of parameters.
We investigate the influence of $\lambda$ on ImageNet Top-1 accuracy, conducting experiments on the ShuffleNetV2 structure.
Table \ref{table:lambda} shows the results.
We find that the optimal $\lambda$ for ShuffleNetV2 \{0.5$\times$, 1$\times$, 1.5$\times$, 2$\times$\} are \{8,4,4,4\}, respectively.
The model capacity has an upper bounded as we increase $\lambda$, and there exists a choice of $\lambda$ to achieve the optimal performance.

%
%

\subsubsection{The influence of $G$.}
To investigate the influence of $G$, we conduct experiments on ImageNet based on ShuffleNetV2.
Talbe \ref{table:group} illustrates the influence of $G$.
We keep $\lambda$ equals to 1, and change $G$ to \{1,2,4\}.
From the result we conclude that increasing $G$ has a positive influence on model capacity. 

\subsubsection{WeightNet on different stages.}
As WeightNet makes the weights for each convolution layer changes dynamically for distinct samples, we investigate the influence of each stage.
We change the static convolutions of each stage to our method respectively as Table \ref{table:stage} shows.
From the result, we conclude that the last stage influences much larger than other stages, and the performance is best when we change the convolutions in the last two stages to our method.

\subsubsection{The number of the global average pooling operator.}
Sharing the global average pooling (GAP) operator contributes to improving the speed of the conditional convolution network.
We compare the following three kinds of usages of GAP operator: using GAP for each layer, sharing GAPs in a block, sharing GAPs in a stage.
We conduct experiments on ShuffleNetV2 \cite{ma2018shufflenet} (0.5$\times$) baseline with WeightNet (2$\times$).
Table \ref{table:gap} illustrates the comparison of these three kinds of usages.
The results indicate that by adding the number of GAPs, the model capacity improves.
 

\subsubsection{Weight similarity among distinct samples.}
\begin{figure}[t]
\begin{center}
 \includegraphics[width=0.5\textwidth]{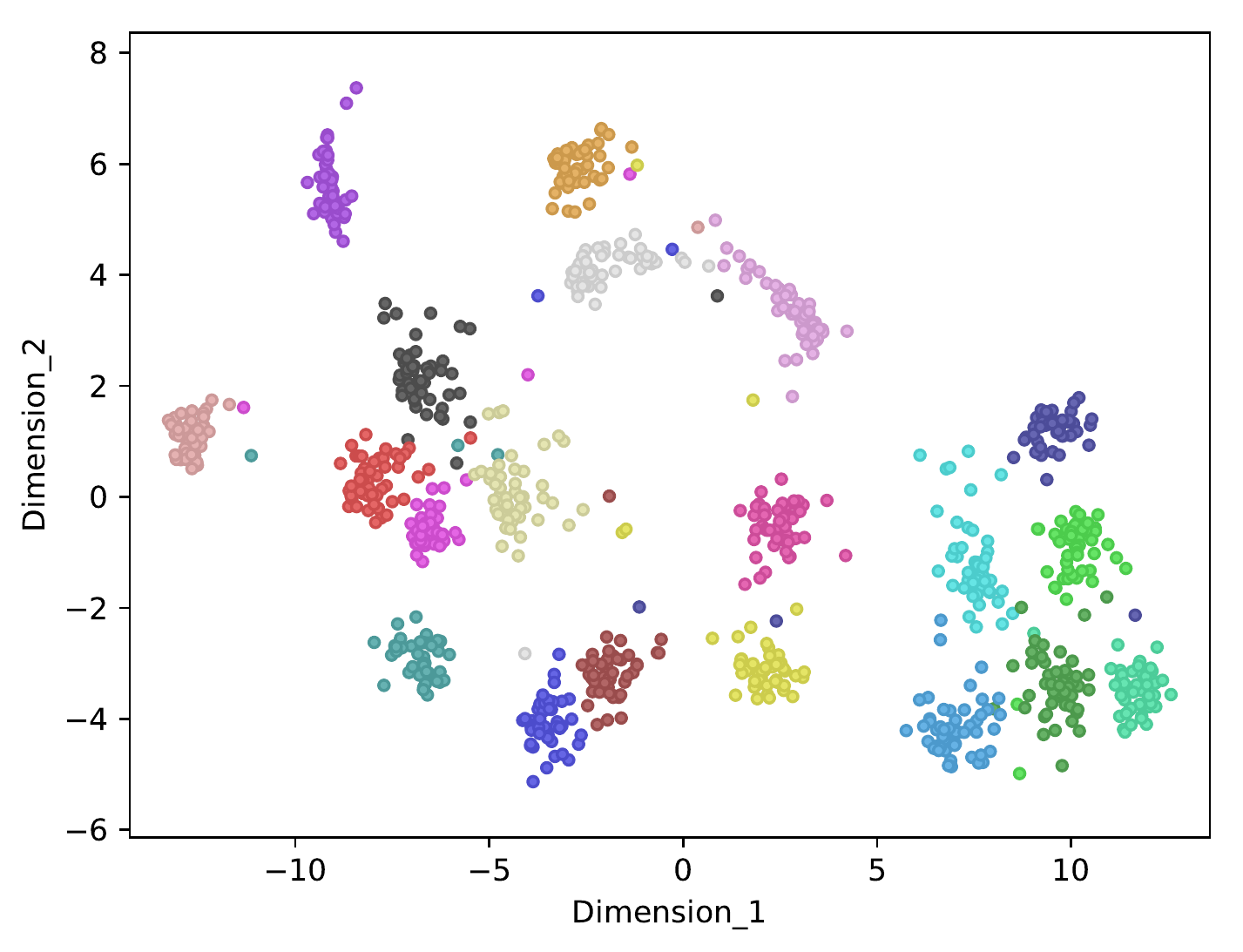}
\end{center}
\caption{Analysis for weights generated by our WeightNet. The figure illustrates of the weights of the 1,000 samples. The samples belong to 20 classes, which are represented by 20 different colors. Each point represents the weights of one sample. }
\label{fig:visual2}
\end{figure}

We randomly select 20 classes in the ImageNet validation set, which has 1,000 classes in total.
Each class has 50 samples, and there are 1,000 samples in total.
We extract the weights in the last convolution layer in Stage 4 from a well-trained ShuffleNetV2 (2$\times$) with our WeightNet.
We project the weights of each sample from a high dimensional space to a 2-dimension space by t-SNE \cite{maaten2008visualizing}, which is shown in Figure \ref{fig:visual2}. We use 20 different colors to distinguish samples from 20 distinct classes.

We observe two characteristics.
First, different samples have distinct weights.
Second, there are roughly 20 point clusters and the weights of samples in the same classes are closer to each other, which indicates that the weights of our method capture more class-specific information than static convolution.

\subsubsection{Channel similarity.}
We conduct the experiments to show the channel similarity of our method, we use the different filters' similarity in a convolution weight to represent the channel similarity.
Lower channel similarity would improve the representative ability of CNN models and improve the channel representative capacities.
Strong evidence was found to show that our method has a lower channel similarity.

We analyze the last convolution layer's kernel in the last stage of ShuffleNetV2 \cite{ma2018shufflenet} (0.5$\times$), where the channel number is 96, thus there are 96 filters in that convolution kernel.
We compute the cosine similarities of the filters pair by pair, that comprise a 96$\times$96 cosine similarity matrix.

\begin{figure}[t!]
\begin{center}
 \includegraphics[width=0.75\textwidth]{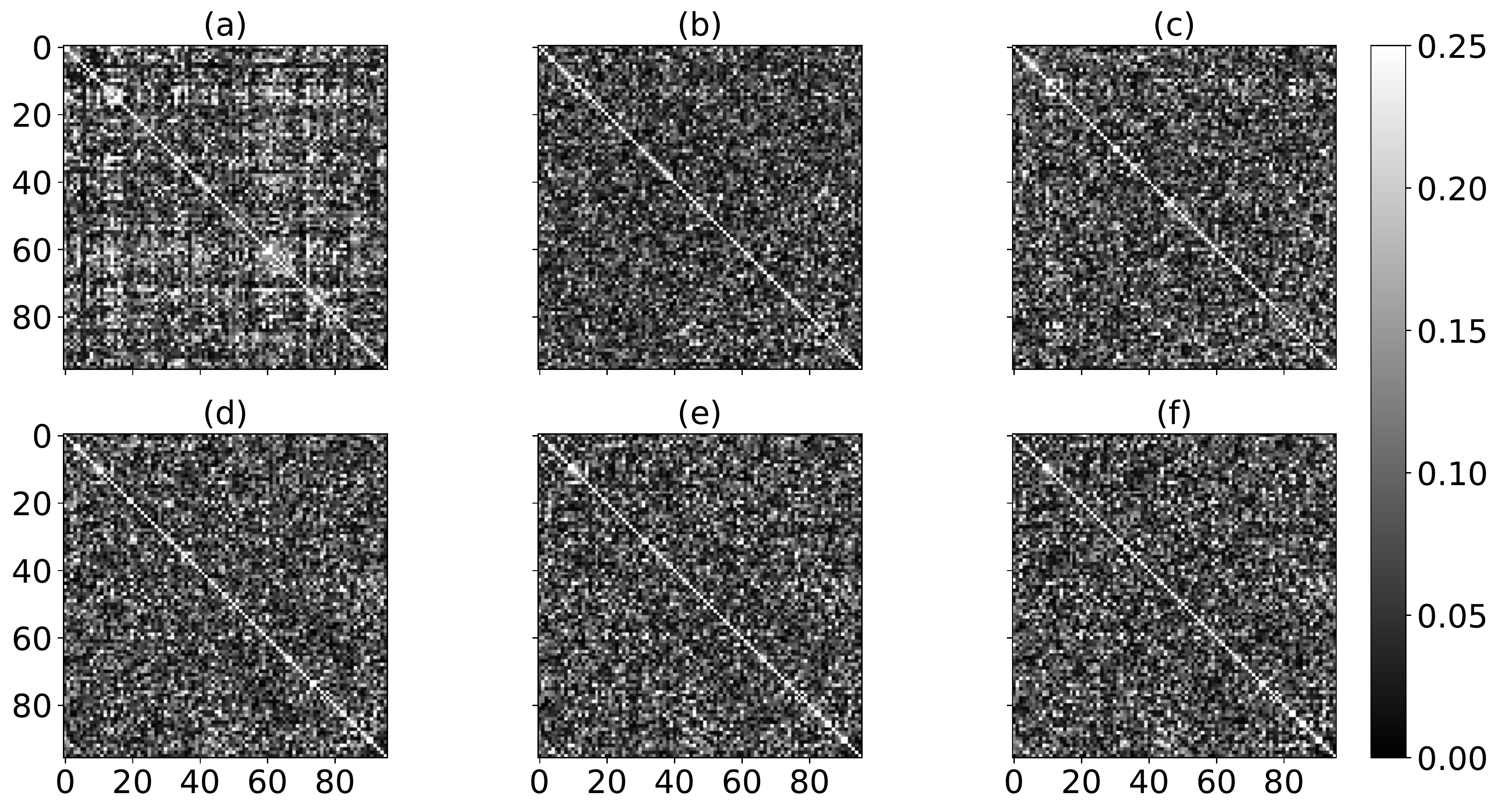}
\end{center}
\caption{\textbf{Cosine similarity matrix.}  
A 96$\times$96 matrix represents 96 filters' pair-by-pair similarity, the smaller value (darker color) means the lower similarities.
(a) Standard convolution kernel's similarity matrix, (b-f) WeightNet kernels' similarity matrixes.
The colors in (b-f) are obviously much darker than (a), meaning lower similarity.}
\label{fig:visual3}
\end{figure}

In Figure \ref{fig:visual3}, we compare the channel similarity of WeightNet and standard convolution. 
We first compute the cosine similarity matrix of a standard convolution kernel and display it in Figure \ref{fig:visual3}-(a). 
Then for our method, because different samples do not share the same kernel, we randomly choose 5 samples in distinct classes from the ImageNet validation set and show the corresponding similarity matrix in Figure \ref{fig:visual3}-(b,c,d,e,f).
The results clearly illustrate that our method has lower channel similarity.

\section{Conclusion and Future Works}
The study connects two distinct but extremely effective methods SENet and CondConv on weight space, and unifies them into the same framework we call WeightNet.
In the simple WeightNet comprised entirely of (grouped) fully-connected layers, the grouping manners of SENet and CondConv are two extreme cases, thus involving two hyperparameters $M$ and $G$ that have not been observed and investigated.
By simply adjusting them, we got a straightforward structure that achieves better tradeoff results.
The more complex structures in the framework have the potential to further improve the performance, and we hope the simple framework in the weight space helps ease future research.
Therefore, this would be a fruitful area for future work.

\subsubsection{Acknowledgements}
This work is supported by The National Key Research and Development Program of China (No. 2017YFA0700800) and Beijing Academy of Artificial Intelligence (BAAI).

\clearpage
%
%
\bibliographystyle{splncs04}
\bibliography{egbib}
\end{document}